# A Comment On *"The Illusion of Thinking"*: Reframing the Reasoning Cliff as an Agentic Gap

Sheraz Khan[1]*, Subha Madhavan[1], Kannan Natarajan[1]
[1] Pfizer Inc., Cambridge, MA, USA
*sheraz.khan@pfizer.com

**ABSTRACT**

The recent work by Shojaee et al. (2025), titled *"The Illusion of Thinking: Understanding the Strengths and Limitations of Reasoning Models via the Lens of Problem Complexity"*, presents a compelling empirical finding, a reasoning cliff, where the performance of Large Reasoning Models (LRMs) collapses beyond a specific complexity threshold, which the authors posit as an intrinsic scaling limitation of Chain-of-Thought (CoT) reasoning. This commentary, while acknowledging the study's methodological rigor, contends that this conclusion is confounded by experimental artifacts. We argue that the observed failure is not evidence of a fundamental cognitive boundary, but rather a predictable outcome of system-level constraints in the static, text-only evaluation paradigm, including tool use restrictions, context window recall issues, the absence of crucial cognitive baselines, inadequate statistical reporting, and output generation limits. We reframe this performance collapse through the lens of an agentic gap, asserting that the models are not failing at reasoning, but at execution within a profoundly restrictive interface. We empirically substantiate this critique by demonstrating a striking reversal. A model, initially declaring a puzzle impossible when confined to text-only generation, now employs agentic tools to not only solve it but also master variations of complexity far beyond the reasoning cliff it previously failed to surmount. Additionally, our empirical analysis of tool-enabled models like o4-mini and GPT-4o reveals a hierarchy of agentic reasoning, from simple procedural execution to complex meta-cognitive self-correction, which has significant implications for how we define and measure machine intelligence. The "illusion of thinking" attributed to LRMs is less a reasoning deficit and more a consequence of an otherwise capable mind lacking the tools for action.

## 1. THE CONTRIBUTION OF THE ILLUSION OF THINKING

"The Illusion of Thinking" represents a significant methodological step forward in the study of machine reasoning. It moves beyond conventional evaluations on static benchmarks, which often suffer from data contamination to a more insightful paradigm (Oren et al., 2023; Deng et al., 2024; Li et al., 2024). Its contributions can be understood through its novel experimental design and the key learnings that emerged.

**1.1 Experimental Design**

The strength of "The Illusion of Thinking" lies in its rigorous experimental framework, designed to probe the limits of reasoning systems systematically (Wei et al., 2022; Jin et al., 2024). The authors have meticulously crafted a testbed that moves beyond conventional, and often contaminated, benchmarks to enable a more controlled and insightful analysis of Large Reasoning Models (LRMs).

**Controllable Environments:** The authors utilize four deterministic puzzle environments: Tower of Hanoi, Checker Jumping, River Crossing, and Blocks World (Fikes & Nilsson, 1971; Ghallab & Traverso, 2004; Russell & Norvig, 2016). These puzzles were chosen because their difficulty can be precisely scaled by adjusting a single complexity parameter, N, which represents elements like the number of disks or blocks. For instance, the Tower of Hanoi's difficulty scales exponentially with N disks ($2^N - 1$ moves), while Checker Jumping scales quadratically ($(N+1)^2 - 1$ moves). This approach facilitates controlled experimentation, a feature often lacking in standard mathematical or coding benchmarks. The puzzles are designed to test core algorithmic reasoning, requiring only the rules provided in the prompt, thereby minimizing reliance on external knowledge. A crucial, though not fully explored, aspect of this experimental design is the rich diversity of computational complexity across the selected puzzles. They do not merely represent an increase in difficulty but probe distinct types of reasoning challenges. This heterogeneity in complexity is a significant strength of the study's design, as it allows for an implicit analysis of how models fail when faced with fundamentally different types of computational problems (Sipser, 2012).

**Model Selection:** The study evaluates several state-of-the-art LRMs, focusing on paired comparisons between "thinking" variants (like Claude 3.7 Sonnet-Thinking and DeepSeek-R1) and their "non-thinking" counterparts (Claude 3.7 Sonnet and DeepSeek-V3). This comparative structure allows for a direct assessment of the "thinking" mechanism's impact. These specific models were selected because they grant access to the intermediate "thinking" tokens, which is crucial for the paper's in-depth analysis. Other models, such as the o3-mini, are included to broaden the analysis of the performance collapse phenomenon across different architectures. For each puzzle instance, the authors generated 25 samples to ensure robust and reliable average performance metrics.

**Trace-Level Validation:** A key innovation is the use of a custom-built puzzle simulators to perform move-by-move adjudication of the models' solutions. This enables an analysis that goes beyond final-answer accuracy to inspect the quality and correctness of the internal reasoning traces, or "thoughts". The simulators are stateful environments that track the configuration of puzzle elements (e.g., disks on pegs) and validate each proposed move against the puzzle's fundamental constraints. This allows for the precise identification of the first point of failure in a sequence of operations, offering a granular view of the models' execution capabilities and limitations.

**1.2 Key Learnings**

This robust design yields several critical and, in some cases, counter-intuitive findings regarding the capabilities of current-generation reasoning models. The principal learnings from the paper are as follows:

1. **A Three-Zone Performance Curve:** The study reveals three distinct performance regimes dependent on problem complexity. At low-complexity, standard, non-thinking models are often more token-efficient and can achieve accuracy comparable to or even greater than their "thinking" counterparts. In the medium-complexity regime, the explicit reasoning of LRMs like Claude 3.7 Sonnet-Thinking provides a distinct and widening performance advantage. At high-complexity, however, all models, both thinking and non-thinking, experience a complete collapse in accuracy to zero.

2. **Pre-Collapse Contraction of Reasoning Effort:** The paper uncovers a paradoxical scaling limit in how LRMs allocate computational effort. As problem complexity increases, the models' reasoning effort, measured by the number of generated "thinking" tokens, initially scales as expected. However, upon approaching the critical complexity point where accuracy collapses, the models counter-intuitively begin to reduce their reasoning effort. This contraction occurs despite the models operating well below their maximum generation budgets, leading the authors to suggest a "fundamental inference time scaling limitation".

3. **Execution as the Limiting Bottleneck:** The findings provide strong evidence that the model's primary limitation is not in devising a solution, but in the faithful execution of its steps. When provided with an explicit, optimal algorithm for the Tower of Hanoi, models like DeepSeek-R1 and Claude 3.7 Sonnet-Thinking still fail at roughly the same complexity point. This indicates that the failure lies in the verification and execution of a long sequence of logical steps, a more fundamental capability than high-level planning.

4. **Complexity-Dependent Inefficiencies in Reasoning:** An analysis of the models' internal reasoning traces reveals shifting patterns of inefficiency. On simpler problems, LRMs often exhibit "overthinking," where they find the correct solution early in the reasoning trace but continue to waste compute by exploring incorrect alternatives. As problems become moderately more complex, this trend reverses, and correct solutions tend to emerge only later in the thought process, after extensive exploration of incorrect paths. At the highest complexities, models fail to find any correct solutions within their reasoning traces.

## 2. THE AGENTIC GAP: FROM EXECUTIONAL FAILURE TO A HIERARCHY OF REASONING

The most revealing finding in Shojaee et al. (2025), and the starting point for our own analysis, is one that isolates the bottleneck of LRM performance: providing models like Claude 3.7 Sonnet-Thinking with an explicit, optimal algorithm for the Tower of Hanoi does not prevent performance collapse. This suggests the bottleneck is not a lack of conceptual knowledge, but a failure of execution. The crucial missing piece of context is that the LRMs are being forced to perform this execution in an environment that is profoundly restricted. The experimental setup forbids the LRMs to write and execute code to solve the puzzles. This prohibition forces the LRM into the role of a "human simulator," painstakingly transcribing thousands of discrete steps, rather than allowing it to function as a "problem solver," which would naturally offload such procedural execution to a more suitable tool (Xi et al., 2025; Qu et al., 2025; Patil et al., 2023).

This exposes the core issue: the observed failures are not of *reasoning in abstracto*, but of agency. Modern LRMs are powerful cognitive engines (Camposampiero et al., 2025; El-Kishky et al., 2025; Kambhampati et al., 2025; Liu et al., 2025; Niu et al., 2024), but the static, text-only interface creates an "agentic gap" (Wang et al., 2024). However, our empirical results show that simply bridging this gap with tools is not sufficient. The way in which a model utilizes those tools reveals a clear hierarchy of reasoning, from simple procedural execution to complex, meta-cognitive self-correction. Drawing from the lexicon of cognitive science, we can frame this distinction as one between *First-Order Agency*, the direct application of a devised strategy to act upon the world, and *Second-Order Agency*, the capacity to reflect upon, evaluate, and revise one's own internal strategies and thought processes.

**2.1 The Baseline: Executional Failure and Learned Helplessness**

Before analyzing the hierarchy of agency in tool-enabled models, it is essential to establish the baseline failure mode that the original study documents. Through a series of controlled interactions, we tasked the o4-mini LRM with solving the "River Crossing" puzzle across several configurations without access to its tool-use capabilities, forcing it to rely on the same text-only, autoregressive reasoning process as the models in the original study.

The results were telling. The model not only failed to produce a valid sequence of moves for any non-trivial case (e.g., $N = 5$ pairs / $k = 3$ person boat) but, after generating extensive yet flawed reasoning traces, repeatedly and confidently concluded in some instances that these solvable problems were "logically impossible." This reveals a critical failure mode: the model's inability to perfectly maintain state and exhaust a complex search space autoregressively leads it to mistake its own executional limitations for fundamental impossibilities of the puzzle itself (mlsubmission, 2025; OpenAI., 2025a). This baseline demonstrates that the non-agentic interface described by Shojaee et al. (2025) is so restrictive that it prevents even the most basic form of problem-solving execution, forcing the model into a state analogous to *learned helplessness*, where the LRM incorrectly generalizes from its inability to act to the conclusion that the task itself is unsolvable (Maier et al., 1976).

**2.2 A Comparative Analysis: The Emergence of a Reasoning Hierarchy**

To probe the nature of reasoning beyond this baseline, a second experimental condition was analyzed, comparing the tool-enabled GPT-4o (a non-reasoning model) and the tool-enabled o4-mini LRM on the same high-complexity task ($N = 20$ pairs / $k = 4$ boat). The outcome, documented in Figure 1, reveals a clear hierarchy of agentic capability, which we can map to established concepts in cognitive psychology.

The GPT-4o model exhibits what we term First-Order Agency, a mode of operation analogous to the fast, intuitive, but error-prone "System 1" of dual-process theory (Kahneman, 2011). After correctly determining that a brute-force search is infeasible, it defaults to a pattern-based heuristic that is plausible but ultimately incorrect (Figure 1A). It then uses its Python tool to execute this flawed plan, generating a complete but invalid solution (Figure 1B). Crucially, it fails to recognize the logical flaw in its own strategy, a flaw that is subsequently exposed by the authors' purpose-built testing simulator (Figure 1C). This process mirrors the human phenomenon of cognitive fixation, specifically the *Einstellung effect*, where a problem-solver becomes rigidly attached to their initial approach, unable to see a more viable alternative even when their current one is failing (Luchins, 1942). GPT-4o successfully uses its tools for execution but lacks the critical cognitive capacity to robustly verify its own strategic output (OpenAI., 2025b). This demonstrates a deficit in the kind of error monitoring and conflict detection that are crucial for complex problem-solving. It possesses the procedural knowledge to act, but not the declarative knowledge about the validity of its own plan, a gap that prevents it from engaging in effective error-driven learning within the task.

In stark contrast, the o4-mini LRM demonstrates Second-Order Agency, a process that more closely resembles the deliberate, analytical "System 2" of dual-process theory (Kahneman, 2011). It too begins with a flawed algorithmic hypothesis (Figure 1D), suggesting that its initial intuitive approach is not infallible. However, its use of the Python tool is more sophisticated; it is not merely for execution, but for verification. The model's own simulation correctly detects the failure of its initial plan. This moment of self-generated negative feedback, a form of cognitive dissonance between its intended outcome and the actual result, triggers a meta-cognitive self-correction, a hallmark of higher-order thinking (Flavell, 1979). The LRM exhibits cognitive flexibility by overcoming its initial fixation, discarding the failed strategy, and selecting an entirely new, correct "*paired-couples*" algorithm (OpenAI., 2025c), which it then successfully validates (Figure 1E) before generating the final solution (Figure 1F). This iterative loop of plan, test, fail, and revise mirrors the process of deliberate practice in human expertise, where feedback is actively sought and used to refine performance (Ericsson et al., 1993). From the perspective of Newell and Simon's problem space theory (1972), GPT-4o performs a limited search and gets trapped on an incorrect path, whereas o4-mini demonstrates a more sophisticated search, capable of backtracking from a failed state to explore a completely different and ultimately *successful branch of the problem space*.

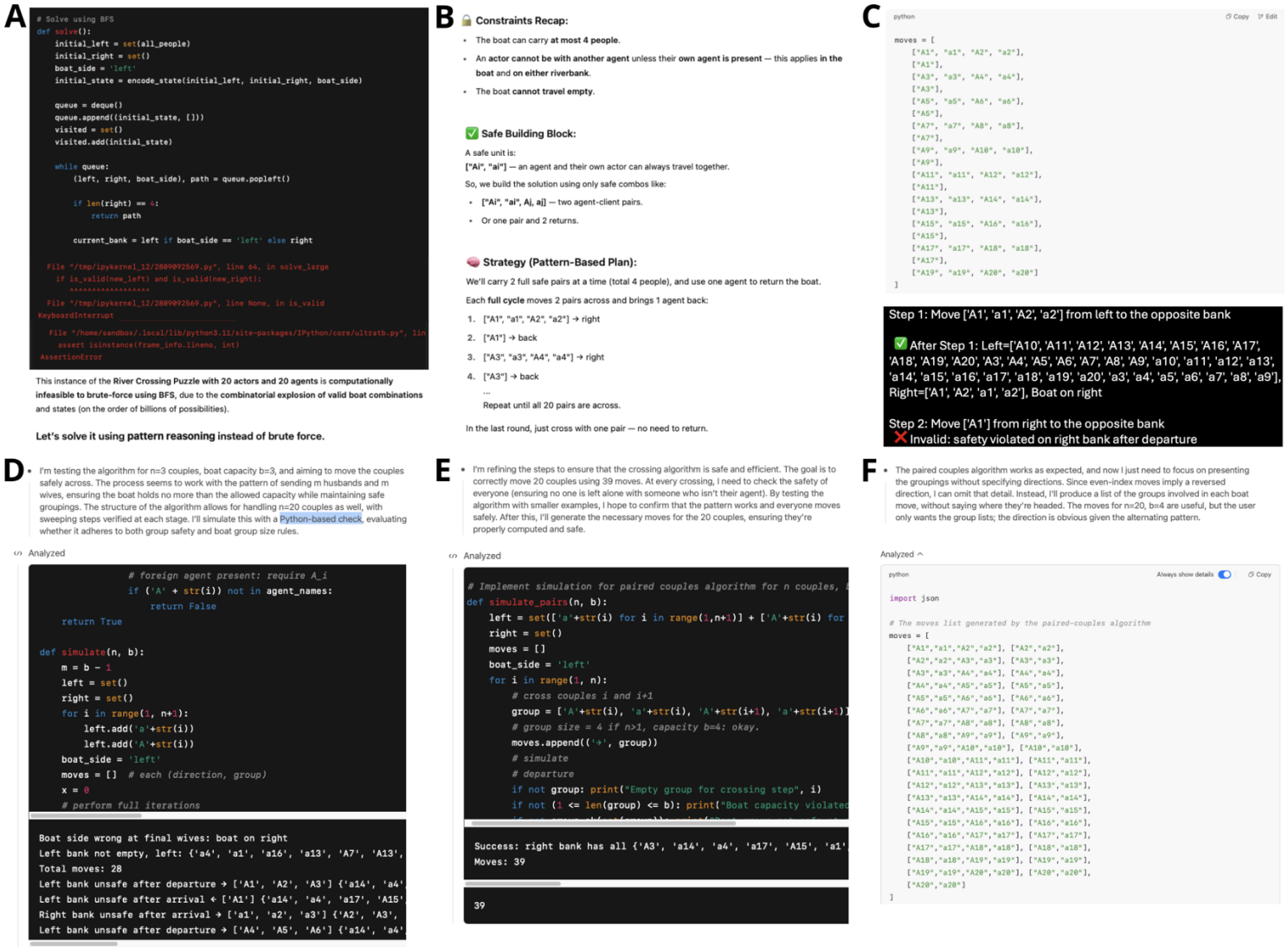

**Figure 1: A Comparative Analysis of Agentic Strategies in a Non-Reasoning Model (GPT-4o) vs. an LRM (o4-Mini), from First-Order Execution to Second-Order Reasoning.** A direct comparison of the problem-solving processes of two tool-enabled models on the high-complexity River Crossing task (*N=20, k=4*). The top row (A-C) illustrates a failure mode in the non-reasoning GPT-4o model analogous to cognitive fixation. **(A)** After determining a brute-force search is infeasible, the model formulates a plausible but flawed heuristic. **(B)** The Model, relying on this flawed heuristic, proceeds to generate a complete but invalid solution. **(C)** The fatal error in the solution is subsequently identified by the authors' purpose-built testing simulator, demonstrating a form of First-Order Agency: the capacity to execute a devised plan but without the meta-cognitive ability to revise the plan itself when it proves flawed. The bottom row (D-F) documents a successful demonstration of meta-cognitive self-correction by the o4-mini LRM. **(D)** LRM begins with a flawed algorithmic hypothesis but uses its simulation tool to detect its own failure. **(E)** Critically, in response to this detected error, the LRM discards its initial strategy and selects a new, correct "*paired-couples*" algorithm, which it then successfully validates. **(F)** The final, correct list of moves is generated. This process exemplifies Second-Order Agency: the ability to monitor, evaluate, and revise one's own problem-solving strategies in response to failure.

This dichotomy provides a powerful, direct validation of our thesis. The tool-enabled GPT-4o succeeded on simpler problems but failed at high complexity because its agency was limited to execution without robust self-correction (OpenAI., 2025b). The tool-enabled o4-mini succeeded because it used its tools to create an interactive feedback loop, allowing it to reason about its own reasoning process (OpenAI., 2025c). This case study demonstrates that the performance cliff documented by Shojaee et al. (2025) is not a boundary of machine reasoning, but a measure of the profound handicap imposed by a non-agentic interface. The true "illusion of thinking" may not be that these models' reason less than we assume, but that the non-agentic paradigm prevents us from observing the full spectrum of their cognitive capabilities.

It is critical to note that the divergence in capabilities between the two models only emerges at a high-complexity frontier. Our analysis of the interaction logs shows that both the GPT-4o and o4-mini models, when granted access to their Python interpreters, successfully solved low to medium complexity variations of the puzzle, $N = 2$, $k = 2$; $N = 3$, $k = 3$; $N = 4$, $k = 3$; $N = 5$, $k = 3$ (OpenAI, 2025b; OpenAI, 2025c). In these cases, both models typically employed a correct Breadth-First Search (BFS) algorithm, demonstrating a shared baseline of agentic competence: they can correctly identify a tractable problem, formulate an appropriate brute-force search, and execute it with a tool to find an optimal solution. The failure point, therefore, is not about the basic ability to use tools, but about the cognitive strategies employed when a problem's complexity renders such exhaustive methods infeasible.

## 3. A RE-EVALUATION OF LRM PERFORMANCE COLLAPSE

While the empirical results of Shojaee et al. (2025) are clear, their interpretation deserves careful examination. The claim of an intrinsic reasoning limit is complicated by several methodological factors that offer a more parsimonious explanation for the models' failure. Our critique of the original study's interface constraints aligns with recent findings by Lawsen (2025) and Twitter thread (@scaling01, 2025), which also identify token limits and puzzle impossibilities as critical confounds. However, our focus moves beyond these methodological issues to the cognitive behaviors they obscure.

### 3.1. Generation Length and Context Window Recall Limitations

A key confounding factor arises from the interaction between task complexity and the models' fixed generation limits. For the Tower of Hanoi puzzle, the solution requires $2^N - 1$ moves. Each move consumes approximately 8 tokens, which is estimated from the discrete tokenization of move elements such as brackets, numbers, and delimiters (e.g., $[1, 0, 2],$). Consequently, the total token cost grows exponentially, as depicted in Figure 2A. This scaling imposes a hard ceiling. For a typical output limit of 64,000 tokens, a performance collapse is not merely possible but mathematically inevitable around $N = 13$, a value corresponding to the failure points reported by Shojaee et al. (2025). The observed failure is therefore better understood not as a "reasoning cliff" but as a predictable "resource cliff," an artifact of the model running out of space to write its answer.

Moreover, this analysis is conservative, as the total generation budget must accommodate not only the final list of moves but also the preceding Chain-of-Thought, or "thinking," tokens. The paper's own data (Shojaee et al., 2025, Figure 6) indicates that this reasoning overhead is non-trivial, consuming 10,000-20,000 tokens for moderately complex tasks. This expenditure on reasoning further reduces the available budget for the final answer, meaning the true resource-based failure point occurs at an N value even smaller than 13. This exacerbates the budgetary constraint, reinforcing the conclusion that the observed performance collapse is an artifact of the experimental design, not a fundamental limitation of the model's reasoning capacity.

A second artifact relates to context window recall limitations (Ding et al., 2024). For puzzles like River Crossing, which require memory of trip history, success depends on maintaining state. As the context window grows, earlier parts of the context containing critical information risk being lost (Liu et al., 2023). This memory loss, inherent to the transformer architecture of all tested models, can directly cause illegal moves and would manifest as a reasoning failure (Kuo, M-T et al., 2023).

### 3.2. Cumulative Error and Statistical Rigor

Beyond resource limits, the paper's interpretation is further complicated by the omission of an analytical baseline for cumulative errors. For any task requiring a long sequence of m operations, the probability of perfect execution, $P_{success} = p^m$, decays geometrically, where p is the per-step success probability. As illustrated in Figure 2B, this decay is precipitous even for highly accurate models. For the Tower of Hanoi with $N = 13$ disks ($m = 8191$ moves), a model with a very high per-move success probability of $p = 0.9999$ still has less than a 45% chance of completing the task perfectly ($0.9999^{8191} \approx 0.44$). The performance collapse reported in the paper may therefore not be a sign of deficient reasoning but may, in fact, align with the expected performance of a robust but imperfect *autoregressive agent*. Without presenting results against this analytical baseline, it is difficult to discern whether a model like Claude 3.7 Sonnet-Thinking is underperforming or behaving precisely as probability theory would predict.

Furthermore, *sample means are presented without confidence intervals or standard errors*. Rigorous statistical validation, such as using bootstrapped standard errors or a mixed-effects analysis, would be necessary to formally establish the significance of reported performance differences. Without such measures of uncertainty, it is difficult to ascertain whether the reported performance gaps, particularly in the medium-complexity regime, represent a true underlying difference in capability or are simply an artifact of sampling variability (Oosterhuis et al., 2024).

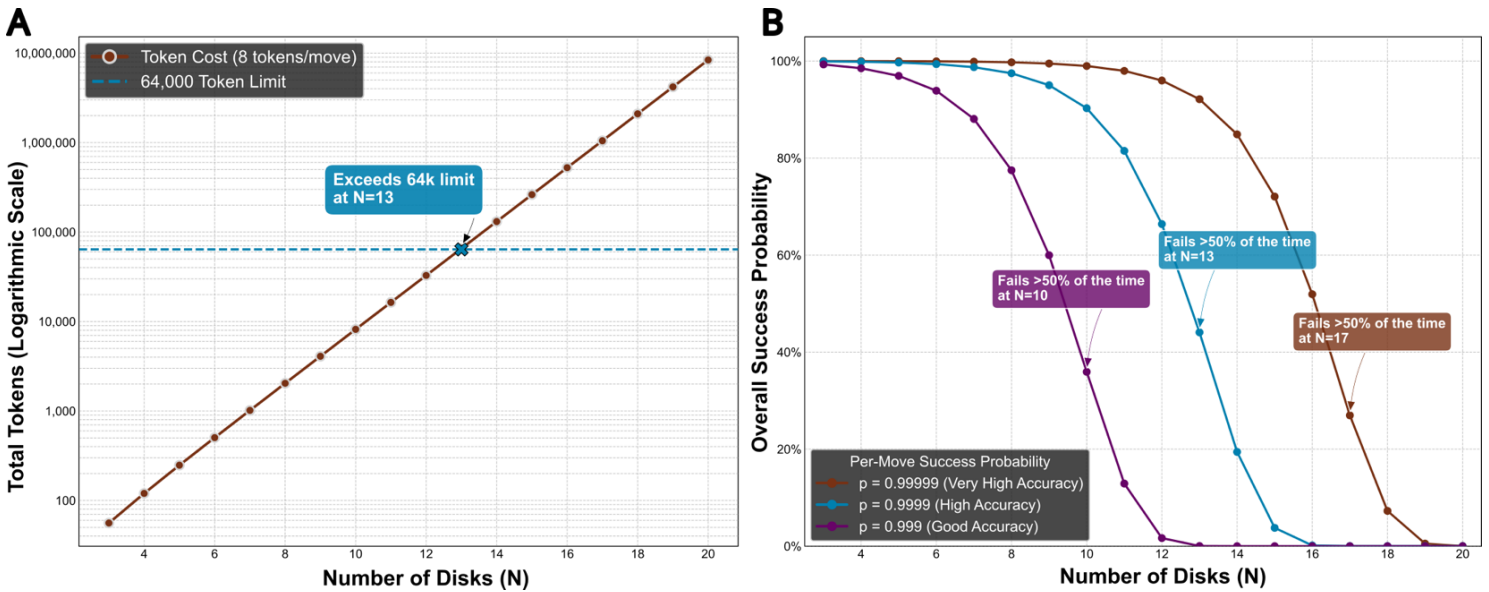

**Figure 2: Predictable Failure Regimes from Resource Limits and Cumulative Error.** The illustration of two critical, resource-based confounding variables that offer an alternative explanation for the performance collapse reported in "The Illusion of Thinking." It frames the collapse as a predictable outcome of system constraints rather than a fundamental limit of reasoning. **(A)** The exponential scaling of token requirements for the Tower of Hanoi solution, plotted on a logarithmic scale. Assuming a cost of approximately 8 tokens per move, the total tokens required for a complete solution ($2^N - 1$ moves) inevitably intersects with the fixed 64,000-token generation limit of the tested models. As annotated, this intersection occurs at N=13, suggesting the observed failure is a predictable "resource cliff." **(B)** The geometric decay of the overall success probability ($P_{success} = p^m$) as a function of the number of disks (N), shown for three different per-move success probabilities (*p*). The plot demonstrates the extreme sensitivity of long-horizon tasks to per-step accuracy. A model with exceptionally high accuracy (p=0.99999) is expected to fail more than 50% of the time only at N=17. In contrast, a model with good, but slightly lower, accuracy (p=0.999) is expected to fail at N=10.

### 3.3. Lack of Human and Cognitive Baselines

The paper's interpretation of the performance collapse is further limited by its comparison set. While comparing models like Claude 3.7 Sonnet-Thinking to their non-thinking counterparts is informative, the study omits critical human and cognitive baselines that prevent a deeper understanding of the observed failure modes.

The paper's discovery of a counter-intuitive scaling limit, wherein models reduce their reasoning effort as complexity approaches a critical point, might not appear counter-intuitive when viewed from a human cognitive perspective. A human problem-solver, when confronted with a task perceived as intractable or unsolvable (e.g., mentally planning thousands of moves for the Tower of Hanoi), would likely not exert maximal cognitive effort in a futile attempt. Instead, they would rationally disengage upon recognizing that the problem's complexity exceeds their cognitive limits (Simon, 1975). This termination of effort is a form of meta-cognitive resource management (Ackerman & Thompson, 2017). The reduction in token usage by o3-mini and Claude 3.7 Sonnet-Thinking might, therefore, be interpreted not as a paradoxical scaling failure, but as an emergent, rudimentary form of this same rational disengagement, potentially learned from patterns where extreme initial complexity correlates with a low probability of success. The absence of these cognitive baselines makes it difficult to attribute the observed failures to a specific limitation of the LRM reasoning versus more universal constraints on complex, sequential problem-solving.

### 3.4. The Confounding Variable of Puzzle Complexity

The paper's primary conclusion of a single "performance cliff" overlooks the critical differences in *why* the models are failing on each puzzle. The analysis tends to treat difficulty as a uniform axis controlled by 'N', but the formal properties of the puzzles suggest the models are succumbing to different cognitive pressures. The early collapse on River Crossing is not just a function of move count but is indicative of a failure to manage the complex, global state constraints inherent to a PSPACE-complete problem. In contrast, failures in Blocks World are more likely tied to the breakdown of strategic, non-linear planning required. Attributing these distinct failures to a single "inference time scaling limitation" is an oversimplification. A more precise conclusion is that the static LRM interface exhibits distinct failure modes when confronted with problems that stress different aspects of *agency*, be it state tracking, strategic planning, reflection, or procedural memory.

## 4. CONCLUSION AND FUTURE DIRECTIONS

"The Illusion of Thinking" provides a valuable service by developing a rigorous benchmark and demonstrating that explicit Chain-of-Thought in models like DeepSeek-R1 and Claude 3.7 Sonnet-Thinking does not guarantee reliable execution of long plans. However, the conclusion of an intrinsic reasoning frontier is premature. The performance collapse documented in the article is significantly influenced by several factors. The true "illusion" might not be the reasoning limitations of these LRMs, but rather the non-agentic paradigm that hinders our observation of their full spectrum of cognitive capabilities. The central question for the field should evolve from a binary "Can models reason?" to a more nuanced "What *kind* of reasoners are they, and under what conditions can they ascend the agentic hierarchy?"

Future research must therefore move beyond simply documenting failures in constrained environments and instead seek to disentangle these variables by adopting a more robust evaluation paradigm. We propose a new standard for benchmarking complex reasoning: LRMs should be evaluated in two distinct modes. The first is the tool-less mode, as performed by Shojaee et al. (2025) which serves as a valuable measure of the procedural friction and brittleness of the underlying autoregressive interface. The second is an agentic mode, where the LRM is granted access to a standard suite of tools, including a code interpreter and a persistent memory scratchpad (Packer et al., 2023). Such a paradigm would allow us to explore several critical research avenues:

- **Probing the Agentic Boundary:** What are the specific task properties or model characteristics that differentiate First-Order from Second-Order agency? Future work should design benchmarks that specifically target meta-cognitive functions like error detection, strategy-switching, and uncertainty estimation.
- **Inducing Higher-Order Agency:** Can models exhibiting only First-Order agency be trained to achieve Second-Order capabilities? This could involve novel training techniques, such as reinforcement learning with meta-cognitive rewards or fine-tuning on datasets that explicitly demonstrate self-correction and strategic adaptation.
- **Architectural Correlates of Agency:** Are there specific LRM's training components or scaling properties that correlate with the emergence of Second-Order agency? Understanding these links is crucial for building more reliable and capable reasoning systems.
- **Implications for AI Safety and Reliability:** A model limited to First-Order agency poses a distinct safety risk: it may confidently execute a flawed or harmful plan without the capacity for self-correction. Fostering Second-Order agency is therefore not just a matter of performance, but a critical step toward building more robust and trustworthy AI.

By embedding LRMs within agentic frameworks and focusing on the cognitive processes they employ, we can more accurately probe the true nature and limits of machine reasoning, distinguishing what these LRMs cannot *think* from what they simply cannot *do* with their hands tied.

**AUTHOR DISCLOSURES:**